# Model Predictive Control Approach to Autonomous Formation Flight

Harun CELIK and Dilara KILINC

*Abstract*— Formation flight is when multiple objects fly together in a coordination. Various automatic control methods have been used for the autonomous execution of formation flight of aerial vehicles. In this paper, the capacity of the model predictive control (MPC) approach in the autonomous execution of formation flight is examined. The MPC is a controller that capable of performing formation flight, maintaining tracking desired trajectory while avoiding collisions between aerial vehicles, and obstacles faced. Through this approach, aerial vehicle models with six degrees of freedom in a three-dimensional environment are performed formation flight autonomously, mostly in a triangle order. Not only the trajectory for the formation flight can be tracked through the MPC architecture, also the collision avoidance strategies of the aerial vehicles can be performed by this architecture. Simulation studies show that MPC has sufficient capability in both cases. Therefore, it is concluded that this method can deal with constraints, avoid obstacles as well as collisions between aerial vehicles. However, implementation of MPC to aerial vehicles in real time holds challenges.

*Keywords*— model predictive control, formation flight, aerial vehicles, collision avoidance

*Authors are with Astronautical Engineering, Erciyes University, Kayseri, Turkiye. Conrresponding author e-mail*: haruncelik@erciyes.edu.tr

# KOL UÇUŞUNUN OTONOM İCRASINDA MODEL ÖNGÖRÜLÜ KONTROL YAKLAŞIMI

Harun ÇELİK[1*]

Dilara KILINÇ[2]

## 1. Giriş

Kol uçuşu, birden fazla varlığın bir düzen içinde birlikte uçmasıdır. Bu varlıkların başında doğada uçan hayvanlar gelir. Birçok teknolojinin geliştirilmesinde esin kaynağı olan hayvan davranışlarından kol uçuşu da insanlar tarafından geliştirilen araçlarla taklit edilir. Bu taklit I. Dünya Savaşı'nda avcı uçakların keşif uçaklarını korumak için görevlendirilmesiyle başlamıştır. İki uçaktan oluşan kol uçuşunun tek uçaktan daha etkili olduğu görülmüş, askeri uçaklar en az iki uçaktan oluşan kollar halinde uçurulmaya çalışılmıştır. Günümüzde de çoğunlukla askeri havacılık ve hava gösterilerinde olmak üzere farklı amaçlarla bu uçuş şekli kullanılır. Ancak bu uçuşu gerçekleştirmek pilotlar açısından oldukça zordur ve büyük dikkat ile kontrol kabiliyeti gerektiren bir görevdir.

Teknolojinin temel hedeflerinden biri insanların hayatını kolaylaştırmak olduğundan mühendisler de pilotların uçuş esnasındaki iş yükünü azaltmak için modern sistemler geliştirmek üzere çalışmalar yapmaktadır. Bu amaçla geliştirilen otonom sistemler uçuşun kontrol bilgisayarı tarafından gerçekleştirilmesi olanağı sunmaktadır. Otopilot olarak da isimlendirebileceğimiz bu sistemler uçuş kararlılığının arttırılmasından tüm uçuş fazlarının otomatik olarak gerçekleştirilmesine kadar farklı görevleri icra edebilen sistemlerdir. Öyle ki bu sistemler pilot hayatını riske atmamak için tamamıyla yer istasyonundan kontrol edilen insansız hava araçlarının geliştirilmesini sağlamışlardır.

İnsansız hava araçları (İHA'ları) otonom olarak kontrol edilebilen, insan gözetimi olmadan uzaktan kontrol edilerek görevler gerçekleştirebilen hava aracı

---

[1] Uzay Mühendisliği Bölümü, Erciyes Üniversitesi, ORCID: 0000-0001-5352-3428
[2] Otonom ve Zeki Sistemler Laboratuvarı, Erciyes Üniversitesi, ORCID: 0009-0001-8638-0440

sistemleridir. İHA'lar gerçekleştirilmesi zor, tehlikeli veya riskli görevleri daha güvenli gerçekleştirebilmeleri nedeniyle birçok alanda kullanılmaktadır. Bu alanlar sınır güvenliği, yangın söndürme, koordineli kurtarma operasyonları, gözetim ve keşif, tarım ve ormancılık gibi hem askeri hem de sivil görevlerde farklı uygulamalardır (Cai vd., 2018; Chen vd., 2022). Tek bir İHA'nın yapabileceği görevler sınırlı olabileceğinden, birbirleriyle etkileşim içinde belirli bir hedefi başarmak için otomatik olarak hareket eden İHA sürüleri daha etkili olabilirler. İş birliği halinde olan İHA'larda, sürü hareketi kavramı hem sürü hareketlerini hem de kolu kapsamaktadır. Kol, gruptaki konumu önceden belirlenmiş bir mekânsal düzene uygun olarak bir araya getirilmiş bireyler topluluğudur. Sürüler ise grup içinde önceden belirlenmiş bir mekânsal düzeni olmayanları da kapsar (Marasco vd., 2012). Sürü İHA'lar sürü hareketinin temel prensiplerini takip ederek yakındaki sürü üyeleriyle çarpışmadan kaçınırlar, diğer sürü üyelerinin hızına uyum sağlamaya çalışırlar ve diğer sürü arkadaşlarına yakın kalmaya çalışırlar (Reynolds, 1987). Kol uçuşları da bu nedenle sürü İHA'ların etkili bir şekilde senkronize olarak karmaşık görevleri daha verimli yerine getirebilmesini sağlayan bir yaklaşımdır.

Uçak ve helikopter gibi pilot taşıyan hava araçlarında otomatik kontrol sistemlerinin kol uçuşunda pilota destek olarak tasarlanması yeterli olabilirken İHA'larda kol uçuşunun tam otonom olarak gerçekleştirilmesi daha elzemdir. Otonom kol uçuşu için hava araçlarının dinamikleri ortak bir kontrol uygulaması aracılığıyla birbirine bağlanır, bu durumda İHA'lar arasında iş birliğinin ve güvenliğin sağlanmasında kol uçuşunun kontrol edilmesinde kritik önemdedir. Etkinliği, kapasitesi ve yetkinliği bir görevin başarısını veya başarısızlığını doğrudan etkiler (Zhang vd., 2020). Kol uçuşu kontrol sistemi, çeşitli bozucu faktörler, arızalar ve belirsizliklerin varlığında bile koordinasyonu, iş birliğini ve güvenliği sağlamayı amaçlamalıdır (Cai vd., 2018; Scharf vd., 2012).

Kol uçuşunun otonom icrası için farklı otomatik kontrol yöntemleri kullanılmıştır. Bunlar arasında PID (Proud, 1999), potansiyel alan yöntemi (Paul vd., 2008; Suzuki vd., 2009), kısıtlama kuvvetleri (Zou vd., 2009), görüntülemeye dayalı yaklaşım (Sattigeri vd., 2004), kayan kipli yaklaşım (Galzi & Shtessel, 2006), uzlaşma temelli yöntemler (Ren & Chen, 2006) yer alır. Ancak bu yöntemler kısıtlamaları ele alırken çeşitli zorluklar içerir (Hu vd., 2012). Kısıtlamalarla başa çıkmakta ise eniyileme temelli yöntemlerin kol kontrol problemleri için daha başarılı oldukları görülmektedir ki bu yöntemlerden en yaygın kullanılanı Model Öngörülü Kontrol (MÖK) yöntemidir (Chao vd., 2011).

MÖK ilk olarak karmaşık kimyasal süreçlerde kullanılmak üzere geliştirilmiş olsa da daha sonra çok değişkenli durumlarla başa çıkma ve optimal girişleri tahmin etme yeteneği nedeniyle hava araçları kontrolünde de hızla kullanılmaya başlanmıştır. Bu bağlamda dört rotorlu ve sabit kanatlı insansız hava araçları MÖK kontrol yöntemi ile istenen bir konumda tutulmuştur (Muresan vd., 2008; Çelik vd., 2016). Bir grup İHA'nın çarpışmasının önlenmesi, araç dinamiklerini stabilize etmek ve dinamik ortamlarda birden fazla sayıda uçan robotlar için yörünge takibi amacıyla kullanılmıştır (Shim vd., 2003). MÖK ile helikopter kontrolü de yapılmıştır (Lopes vd., 2006). Birden fazla İHA kol uçuşunun otonom icrası için de MÖK algoritmaları kullanılmıştır (Shin vd., 2011; Vegas vd., 2002). Motor itki kuvvetleri, açısal ve çizgisel hareketler üzerindeki kısıtlamalar ve çarpışma önleme kısıtlamaları dikkate alınarak kapsamlı kol uçuşu kararlılığı ve otonom seyrüseferi için de kullanılan MÖK (Bemporad & Rocchi, 2011), uzlaşma tabanlı kontrol yöntemiyle de uygulanmıştır (Kuriki & Namerikawa, 2015). Yapılan bu çalışmalar kol uçuşunun otonom icrasında MÖK yaklaşımının iyi bir kapasiteye sahip olduğunu ortaya koymaktadır. Bu bölümde de kol uçuşu, otomatik kontrol ve MÖK ele alınarak bahsi geçen kapasitenin daha iyi açıklığa kavuşturulması hedeflenmektedir.

**2. Kol Uçuşu**

Kol uçuşunun hava araçlarında farklı performans üstünlükleri sağladığı görülmektedir. Bunlardan biri kol uçuşunda kanat adamına etkiyen sürüklemenin azalmasıdır. Kol uçuşunda arkadaki araçlar, önde giden araçların kanatlarını terk eden hava akışlarının etkisiyle havada kalmak için gerekli olan taşıma kuvveti ile indüklenen sürüklenmeyi azaltırlar. Sürüklemenin azalmasıyla kanat adamlarının yakıt sarfiyatı azaltılabilir. Diğer taraftan hava araçlarının kol oluşturarak birbirlerine çok yakın uçmaları sayesinde radar sistemleri bu araçları tek bir araç olarak algılayabilir (Çelik vd., 2021). Böylece radardan kaçınmak için de kol uçuşu icra edilebilir. Ayrıca tanker uçaklardan yakıt ikmali yapabilmek için de kol uçuşu icra edilir (Çelik vd., 2022). Dünyada ilk defa bir taarruz İHA'sı ile insansız savaş uçağının tam otonom bir şekilde kol uçuşu icra ettiği sırada alınan bir görüntü Şekil 1'de verilmiştir.

Kol uçuşunu icra eden kontrol sistemleri farklı yapısal yaklaşımlarla tasarlanabilir. Bu yaklaşımlar arasında lider-takipçi, davranışsal ve sanal yapılar bulunmaktadır. Lider-takipçi tabanlı mimari en sık kullanılan yapıdır (Das vd., 2002; Sattigeri vd., 2004a; Scharf vd., 2003). Bu yaklaşımda, bir araç kol uçuşu için izlenecek güzergâhı sağlayan lider olarak belirlenir. Diğer araçlar kanat adamı olarak adlandırılırlar ve takipçi olarak belirlenirler. Davranışsal mimaride, kolda yer alan tüm üyeler kendi belirlediği davranışlarıyla çevrelerini ve komşu üyeleri kontrol etmeyi hedeflerler. Son kontrol kararı ise her aracın kendi davranışının göreli ağırlığına dayalı olarak alınır (Scharf vd.,2011; Balch & Arkin, 1998). Bir diğer kol yaklaşımı da sanal bir yapıdır. Burada hava araçları sanal olarak belirlenen bir hareketli noktayı takip ederler ve tüm araçlar tek bir sanal yapıya gömülü bir şekilde davranır (Leonard & Fiorelli, 2001; Ren & Beard, 2004).

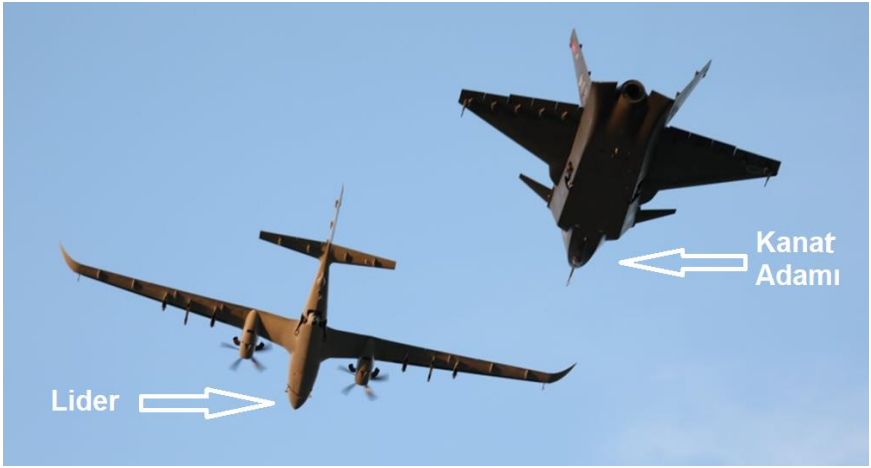

*Şekil 1* Bayraktar Akıncı (lider) ve Kızılelma'nın (takipçi/kanat adamı) kol uçuşu

Sanal yapıyla sahip kol uçuşu mimarisi, sağlamlık ve lider bağımlılığı ile ilgili sorunların etkisini azaltmayı amaçlar. Bir liderin belirlediği güzergâhı takip etmek yerine, her araç sanal olarak hareket eden bir noktayı takip eder. Daha önce merkezi bir bilgisayar tarafından oluşturulan hareketli referans noktaları olan bir veya daha fazla sanal nokta/lider olabilir (Richards & How, 2005; Shim vd., 2003). Koldaki tüm araçlar tek bir nokta olarak alınabildiğinden sanal yapı mimarisi kullanılarak bir kolun yönlendirilmesi, diğer yaklaşımları kullanmaktan daha kolaydır. Bununla birlikte, koldaki hava araçları manevraları sadece birlikte ve aynı şekilde yapabilir. Bu şekilde diziliş şekli korunurken engellerden kaçınmak için koldaki araçların farklı manevralar yapmaları gerektiği durumlarda zorluklar yaşanır (Shin vd., 2011).

Lider-takipçi kol kontrolü yapısı ise diğer kol kontrol yapılarına kıyasla, istenen kol düzenini daha etkin koruyabilir ve karmaşık görevleri daha etkili bir biçimde başarabilir (Wu vd., 2022). Bu mimari koldaki her aracın otomatik kontrolcülerinin hiyerarşik bir düzenlemesini, en az biri lider (Muresan vd., 2011) olacak ve geri kalanlar takipçi olacak şekilde ayarlamaktan oluşur. Alanyazında lider-takipçi yapısının da birçok türü bulunmaktadır. Bunlar arasında İHA'ların bir zincirde uçması, birden fazla liderin olması veya ağaç benzeri topolojilerin kullanılması gibi yöntemler yer almaktadır. Lider-takipçi kollarında hareketi belirlemek için liderin hareketine karar verilmesi yeterli olabilir ancak bu da kolun tamamı için tek bir başarısızlık noktası oluşturur. Ayrıca lider için kanat adamlarından geri besleme mekanizması olmadığı için bir takipçinin lideri yakalayamaması durumunda sorunlar ortaya çıkabilmektedir (Fahimi, 2008). Bu sorunların üstesinden gelebilmek için uçuş planlama ve güzergâh takibi tüm araçlar için oluşturulabilir. Bu şekilde hava araçlarının bir görevin başlangıcından sonuna kadar engellerden kaçınması, net bir güzergâh belirlenmesi, çarpışmalarının önlenmesi, koldaki bir veya daha fazla üyenin başarısızlığı durumunda güvenlik sağlanabilmektedir (Hejase vd., 2015).

Davranışsal mimaride ise referans alınan merkezi bir hava aracı bulunmamaktadır. Merkezi olmayan bir kol uçuşunda herhangi bir lider olmaksızın birlikte hareket eden araçlar mevcuttur. Her bireyin durum vektörü ile ilgili bilgi diğer araçlara da iletilir. Lider-takipçi yapısındaki hiyerarşik ağaç şeklindeki ara bağlantılardan farklı olarak döngü içeren bir yapıya sahiptir. Araçlar arasında karşılıklı bir bağımlılık ilişkisi vardır, yani kontrolcülerin düzeni sıralamaya tabi değildir (Chao vd., 2011). Bu mimari daha çok büyük kollar ve sürüler için uygundur. Bu mimaride, sadece bir liderin güzergâh bilgisinin değil, tüm araçların bilgisinin dikkate alındığı, lidere bağlı kolların aksine işbirlikçi bir davranış vardır (Lopes vd., 2006).

### 3. Model Öngörülü Kontrol

Model öngörülü kontrol (MÖK), belirli bir tek kontrol stratejisini ifade etmekten ziyade, sistem modelini kullanarak bir amaç fonksiyonunu minimize etmeyi hedefleyen ve kontrol sinyalini oluşturan çeşitli kontrol yapılarını ifade eden genel bir terimidir (Özdemir, 2015). Model Öngörülü Kontrol yöntemi ilk olarak güç reaktörlerindeki ve petrol rafinerilerindeki kontrol gereksinimlerine bağlı olarak özel geliştirilmiş, günümüzde kimya, gıda, otomotiv, havacılık, metalürji ve kâğıt endüstrilerini kapsayan geniş bir alana da uygulanmaktadır (Danayiyen, 2013).

MÖK'ün temel amacı sistem davranışını tahmin etmek için bir dinamik model kullanmak ve tahmini optimize ederek en iyi kararı -yani kontrol hareketini- üretmektir (Özdemir, 2015). MÖK ile yapılan kontrolün temel hedefleri (Qin & Badgwell, 2003):

- Girdi ve çıktı kısıtlarının aşılmasını önlemek,
- Bazı çıktı değişkenlerinin en iyi değerlerini bulurken diğer çıktıları belirli aralıklarda tutmak,
- Girdi değişkenlerinin gereksiz sapmalarını engellemek,

- Gelecekteki sistem davranışını tahmin etmek ve bu tahminleri kullanarak daha iyi bir kontrol stratejisi oluşturmak

şeklinde verilebilir. Bu hedefleri birlikte başarmaya çalışan MÖK, genel olarak sistemin performansını iyileştirmeyi, istenen hedeflere ulaşmayı ve güvenliği sağlamayı amaçlayan bir kontrol yöntemi olmaktadır.

### 3.1. Çalışma Prensibi

MÖK hesaplamaları, bir sistemin mevcut/ölçülmüş çıkış/yanıt değerleri ile bu çıkışların gelecekteki değerlerinin öngörülerine dayanır. MÖK kontrol hesaplamalarının amacı, öngörülen yanıtın istenen/referans değer noktasına en iyi şekilde ulaşması için kontrol girişlerinin sırasını belirlemektir. Bu en iyi yaklaşma hesabı kontrolün eniyilemeye dayanmasına neden olur. Eniyileme ile hedeflenen ise sistem dinamikleri ile girişlerin belirli sınırları (kısıtlar) dahilinde sonlu ufuklu bir optimal kontrol probleminin çözülmesidir. Ele alınan zaman diliminin ufuk olarak adlandırıldığı bu yöntemde bir sürecin veya sistemin modeli kullanılarak belirli bir öngörü ufku için her $t$ zamanında gelecekteki çıkışlar tahmin edilir. Bir kontrol ufku dahilinde sisteme uygulanması öngörülen girişler belirli bir eniyileme kriterine göre hesaplanır. Hesaplanan bu en uygun girişlerden ilki sisteme uygulanarak eniyileme problemi yeniden çözülerek sonraki adım için öngörülen giriş değerleri elde edilir. Bu şekilde devam eden işlemler ve değişkenler Şekil 2 ile gösterilebilir.

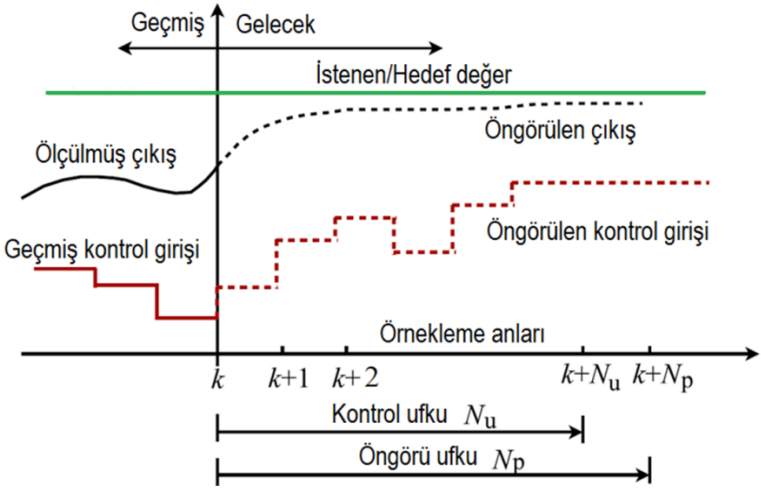

*Şekil 2* Model öngörülü kontrol örneği

Ayrık zamanlı doğrusal zamanla değişmeyen bir sistemin durum uzay modeli

$$x(t+1) = Ax(t) + Bu(t) \tag{1}$$

$$y(t) = Cx(t) \tag{2}$$

olarak ve her $t \geq 0$ değeri için de

$$y_{\min} \leq y(t) \leq y_{\max}, \; u_{\min} \leq u(t) \leq u_{\max} \tag{3}$$

kısıtlarını sağlayacak; $x(t) \in \mathbb{R}^n$ durum, $u(t) \in \mathbb{R}^m$ kontrol ve $y(t) \in \mathbb{R}^p$ çıkış değerlerini içeren matrisler olacak şekilde verilecek olursa bu sistemde kontrol ufku dahilinde girişlerin öngörüsü eniyileme probleminin çözümüne dayalı olarak her $t$ zamanı için

$$\mathbf{U} \triangleq \min_{(u_t, \ldots, u_{t+N_u-1})} \left\{ J(\mathbf{U}, x(t), N_p, N_u) = x_{t+N_p|t}^T P x_{t+N_p|t} \ldots \right. \\ \left. + \sum_{k=0}^{N_p-1} \left[ x_{t+k|t}^T Q x_{t+k|t} \right] + \sum_{k=0}^{N_u-1} \left[ u_{t+k|t}^T R u_{t+k|t} \right] \right\} \tag{4}$$

ifadesiyle ve

$$y_{\min} \leq y_{t+k|t} \leq y_{\max}, \; k=1,\ldots,N_c,$$
$$u_{\min} \leq u_{t+k} \leq u_{\max}, \; k=1,\ldots,N_c,$$
$$x_{t|t} = x(t),$$
$$x_{t+k+1|t} = A x_{t+k|t} + B u_{t+k}, \; k \geq 0,$$
$$y_{t+k|t} = C x_{t+k|t}, \; k \geq 0,$$
$$u_{t+k} = K x_{t+k|t}, \; N_u \leq k < N_p,$$

şartlarını sağlayacak şekilde bir maliyet fonksiyonu ile hesaplanabilir. Burada $x_{t+k|t}$, $x(t)$ durumundan başlanarak (2) bağıntısına $u_t, \ldots, u_{t+k-1}$ kontrol girişlerinin uygulanmasıyla elde edilen $t+k$ zamanında öngörülen durum değişkenleri ifade etmektedir. (4) bağıntısında ise $N_u \leq N_p$ ve $N_c \leq N_p$ olacak şekilde $N_p$ çıkış öngörü ufkunun, $N_u$ kontrol girişi ufkunun ve $N_c$ kısıt ufkunun uzunluğudur. Ayrıca $P$ durum ağırlık matrisi, $Q$ giriş ağırlık matrisi ve $R$ belirli bir zaman ufku kontrol probleminde çözüm kararlılığı için önemli olan (Chao vd., 2011) son durum ağırlık matrisidir. Bu sayede (4) bağıntısındaki eniyileme probleminin *t* anında en iyi çözümü

$$\mathbf{U}^*(t) \triangleq \left\{ u_t^*, \ldots, u_{t+N_u-1}^* \right\}$$

olur. Çözülerek öngörülen kontrol girişlerinden sadece ilki olan $u_t^*$ (2) bağıntısıyla verilen sistem çıkışına uygulanarak yeni kontrol girişleri elde edilir ve bu işlemler her adımda tekrarlanır. Süreç, her yinelemede bir zaman adımı ilerleyerek, tahmin ve kontrol ufkunu kaydırarak sürekli olarak uygulanır. MÖK anlık olarak optimal bir kontrol problemini çözer, gelecekteki kontrol girişlerinin sırasına ve sistem davranışına dayalı olarak kontrol girişlerini elde eder, kontrolü tahmin ufku içinde verilen hedefe göre uyarla, yani sistem çıkışının istenen değere en iyi şekilde yaklaşmasını sağlar.

MÖK'ün ayırt edici özelliği buradan da görüldüğü üzere hareket eden ufuk yaklaşımıdır. Bu yaklaşıma göre yukarıda matematiksel olarak verilen kontrol süreci şu adımlarla gerçekleştirilir (Camacho, 2007).:

1. Her zaman adımında (*t*), gelecekteki çıkış değerleri bir öngörü ufku boyunca hesaplanır. Bu öngörülen çıkışlar, geçmiş giriş ve çıkış değerlerine ve gelecekteki kontrol işaretlerine bağlı olarak elde edilir.

2. Gelecekteki kontrol işaretleri, sürecin istenen/hedef değerine mümkün olduğunca yakınsamasını sağlamak için bir eniyileme yöntemi kullanılarak hesaplanır. (4) bağıntısında olduğu gibi genellikle bu eniyileme yöntemi, öngörülen çıkışlar ile öngörülen hedef değer arasındaki farkın ikinci dereceden bir fonksiyonu şeklinde tanımlanır. Aynı zamanda kontrol artışları da genellikle bu eniyileme yöntemi içerisinde yer alır.

3. Elde edilen kontrol işaretleri arasından sadece birinci kontrol işareti sisteme iletilir, çünkü bir sonraki örnekleme anında bir sonraki çıkış değeri ölçülebilir hale gelecektir. Bu nedenle, yeni bir kontrol işareti hesaplaması yapılması gerekecektir. Bu durum hareket eden ufuk kavramı olarak bilinir.

Anlaşıldığı üzere MÖK, kontrol sürecinde en iyi performansı elde etmek için öngörülerin, eniyileme yöntemlerinin ve hareketli ufuk kavramının birleşimini sağlayarak bunları birlikte kullanır. Bu sayede MÖK'ün, parametre ayarlamalarının diğer yöntemlere göre daha kolay olması, geniş bir uygulama alanı, çok değişkenli sistemlerin kontrolünde kullanılabilmesi, sınırlamaların üstesinden gelmenin tasarım aşamasında kolay olması gibi üstünlükleri bulunur. Ancak MÖK'ün kontrol sistemlerinde kullanılırken uygun bir sistem modelinin elde edilmesinin zorluğuna, gerçek sistemle model arasındaki farklılık arttıkça istenilen sistem yanıtını elde etmenin zorlaştığına, sistem dinamiği değişmediği sürece kontrolcü önceden elde edilebilirken uyarlamalı kontrol yapılıyorsa tüm hesaplamaların her örnekleme zamanında tekrarlanmasının gerektiğine de dikkat edilmelidir. Bu nedenle performansı etkileyen kontrol değişkenleri iyi ayarlanmalıdır.

### 3.2. Kontrol Değişkenleri

Model öngörülü kontrol değişkenleri, kontrol algoritmasının performansını etkileyen ve ayarlanması gereken çeşitli bileşenleri içerir. Ayarlanması gereken bu parametreler öngörü ufku, kontrol ufku, kısıtlar ve ağırlıklardır.

*Öngörü ufku*, MÖK algoritmasının gelecekteki bir zaman aralığında öngörüler yapacağı süreyi belirler. Daha uzun bir öngörü ufku, sistemin gelecekteki davranışını daha iyi tahmin etmesini sağlar, ancak uzun seçilmesi hesaplama süresini ve hesaplama karmaşıklığını artırabilir. Öngörü ufku, sistem dinamiği, hedeflere ulaşma süresi ve diğer performans gereksinimleri göz önünde bulundurularak belirlenmelidir.

*Kontrol ufku*, MÖK algoritmasının eniyilenecek kontrol girişi değerlerinin sayısını belirler. Daha uzun bir kontrol ufku, daha fazla kontrol hareketinin planlanmasını sağlar, ancak uzadıkça hesaplama süresini ve hesaplama karmaşıklığını artırır. Kontrol ufku, sistem hızı, dinamik tepkiler ve uygulanabilirlik gereksinimleri gibi faktörlere bağlı olarak seçilmelidir.

MÖK algoritmasının eniyileme yaparken dikkate alması gereken sistem sınırlıklarını *kısıtlar* ifade eder. Bu kısıtlar, giriş ve çıkış değişkenleri üzerindeki sınırlamaları temsil edebilir. Örneğin, giriş değişkenlerinin belirli bir aralıkta kalması veya çıkış değişkenlerinin belirli bir değeri aşmaması gibi kısıtlar MÖK algoritması tarafından gözetilir.

*Ağırlıklar* ise kontrol algoritmasının eniyileme sürecinde farklı hedefler arasındaki önceliği belirlemek için kullanılır. Örneğin, çıkış hedeflerinin hedef değerlere ne kadar yakın olması gerektiği veya kontrol hareketlerinin ne kadar küçük veya pürüzsüz olması gerektiği gibi hedefler arasında bir denge kurulmasını sağlar. Ağırlıklar, kontrol performansı ve istenen davranışa bağlı olarak ayarlanmalıdır.

### 4. Otomatik Kol Uçuşu İçin Model Öngörülü Kontrol

Otomatik kol uçuşunda uygun güzergâhın oluşturulması ve takip edilmesi, koldaki hava araçlarının birbirlerine göre istenen konumda tutulmaları, koldaki hava araçlarının birbirleriyle veya karşılaşılabilecek engel ve diğer hava araçlarıyla çarpışmalarının önlenmesi gibi temel görevler yer alır. Otomatik kol uçuşunun temel amacı; belirli bir referansı takip ederken ve koldaki hava araçlarının genel davranışını kontrol ederken, birlikte hareket şeklini de korumaktır. Otomatik kol uçuşunda istenen güzergâhın takibi ve gerekli görevler tam otomatik olarak gerçekleştirilebilir.

Kol uçuşunu otonom olarak icra etmek için birçok kontrol yöntemi uygulanmaktadır. Bu yöntemler merkezi, merkezi olmayan veya hiyerarşik kontrol yöntemi olarak üç sınıfta toplanabilir. Kontrol yöntemleri seçilirken iletişim yetenekleri, sağlamlık ve hesaplama yetenekleri gibi faktörler dikkate alınır. Günümüzde kavramsal tasarımlarda ve uygulamada yaygın olarak kullanılan kontrol yöntemleri merkezi olmayan ile hiyerarşik kontrol tasarımlarıdır. Model öngörülü kontrol ise merkezi ve merkezi olmayan yaklaşımla tasarlanabilir.

Merkezi MÖK yöntemi tüm hesaplamaların ve işlemlerin merkezi bir kontrol ünitesinde gerçekleştirilmesi nedeniyle merkezi olarak adlandırılır. Bu kontrol ünitesi, kolun tümünü koordine eder ve ortak bir kontrol stratejisi oluştururken, her bir hava aracının hareketini ve kontrol girişlerini eniyiler. Ancak kontrol bilgilerinin merkezi olarak işlenmesi, yüksek hesaplama gücüne ve geniş iletişim altyapısına ihtiyaç duyulması anlamına gelir. Bu durum kolun iletişiminde gecikmelere ve işlem yükünün artmasına neden olur. Bunlar da kolun performansının düşmesine, hata ve başarısızlık ihtimalinin artmasına neden olur (Mengali vd., 2009).

Merkezi olmayan MÖK yönteminde denetim eylemlerinin hesaplanması ve işlemlerinin gerçekleştirilmesi her bir hava aracının kendi işlemcilerine dağıtılarak yapılır. Hesaplamaların dağıtılması hava araçlarının koldaki diğer tüm araçlarla iletişim kurmak yerine, sadece yakınındaki birkaçına göre kontrol bilgilerinin işlenmesinin yeterli olması anlamına gelmektedir. Bu yöntemde her bir hava aracı için oluşturulan komutların çakışma ihtimali azalmakta ve kolun kararlılığı açısından üstünlük sağlanmaktadır (Keviczky vd., 2004).

Otomatik kontrol yapılırken işlenmesi gereken veri hacmi arttıkça yapılması gereken hesaplamalar artacağından MÖK'ün tasarımında tercih edilecek yaklaşım seçilirken hesaplama gereksinimleri dikkate alınmalıdır. Nitekim hesaplamaların uzaması nedeniyle geç üretilecek bir denetim komutunun zamanı geçince bir işe yaramaz. Hesaplama yükünün ise sanal ve davranışsal kol uçuşu mimarisine nispeten lider-takipçi mimarisinde daha düşük olduğu söylenebilir. Nitekim farklı hava araçları dinamikleri ile bunların birbirleriyle olan etkileşimleri de modellenerek birlikte kontrol edilmeye çalışılması hesaplama yükünü arttırmaktadır. Bunun yerine koldaki hava araçlarının aynı olduğu, tek bir hava aracı dinamik modelinin diğer hava araçlarının dinamik davranışını yansıttığını, maliyet fonksiyonunun ve kısıtlamaların koldaki kanat adamları için aynı olduğu bir kontrol amacıyla eniyilemenin yapılması işlem hacmini azaltmaktadır.

Farklı modeller yerine tek bir model alınsa da uçak dinamiği doğrusal olmayan birçok değişken içermektedir. Böyle bir dinamiğe sahip aracı kol uçuşu için kontrol ederken doğrusal olmayan dinamik kısıtlamalarına ve eyleyici limit kısıtlamalarına uyarak kontrol girişlerinin üretilmesi gerekir (Kuriki & Namerikawa, 2015). Tasarlanan kol uçuşu kontrol yöntemi, bu kısıtlar altında sıkı bir kol uçuşu doğruluğu sağlamalıdır.

Lider takipçi yapısındaki kol uçuşunda kanat adamı konumunun liderin konumuna göre sabit tutulması istenir. Sıkı bir düzen için kontrol yasaları geliştirilirken lider hava aracının belirlenmiş bir uçuş yolu olur ve kanat adamının diziliş şekliyle aerodinamik faydalar sağlaması amaçlanabilir. En basit modelde lider uçak, kolun hava akışından etkilenmez, yalnızca kanat adamı için kontrol sisteminin tasarlanması gerekir. Bu sayede sadece kanat adamın doğrusal olmayan dinamikleri için modelleme yapılabilir. Bu gibi modellerde, yakıt tasarrufu elde etmek için de kanat adamı, maksimum aerodinamik fayda için lider uçağa göre belirli bir konumu takip eder (Bemporad & Rocchi, 2011). Lider sadece kendi uçuş düzlemini belirleyip o doğrultuda uçmaya çalışırken takipçinin hareketlerini çoğu zaman dikkate almaz. Takipçi ise sadece liderin hareketini izlemeyi ve onunla kendi arasında belirlenen konumda kalmayı başarmaya çalışır.

Belirli bir uçağın lider olarak seçilmesi yerine uçaklar arasında sanal bir nokta oluşturularak kolun bu noktayla uçarak güzergâhı takip etmeleri sağlanabilir. MÖK, her hava aracı için kendisinin ve koldaki diğerlerinin verileri de dahil olmak üzere hava aracının sanal noktaya göre kullanabileceği verilere dayalı olarak optimum girdiyi bulan bir kontrolör halini alır. Dolayısıyla merkezi bir kontrol yerine sanal yapı oluşumunu kontrol etmek için sanal lideri takip problemini çözmek gerekir.

Bir liderin veya sanal bir noktanın takibinde kontrol sisteminin çıkışı ve hedefi, kolu istenen konumda tutmak için gerekli bilgiler olan üç boyutlu konum verilerinden oluşur. Şekil 3'te gösterildiği gibi, sanal noktanın veya lider uçağın konumu ve yönelimine göre her bir kanat adamının hedef uçuş düzlemi doğru ve tutarlı bir şekilde takip etmesini kontrol sistemi sağlar.

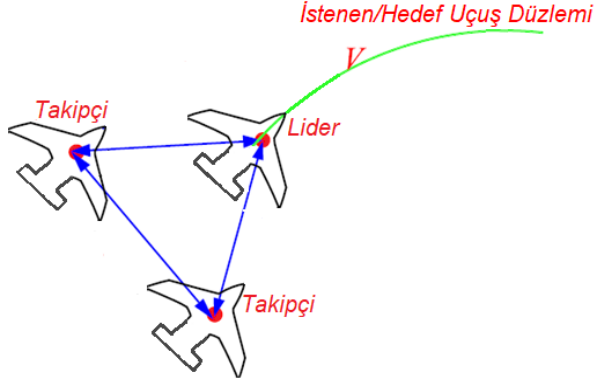

**Şekil 3** *Üçgen düzeninde kol uçuşu*

Kol uçuşunun otomatik olarak icra edilebilmesi için başarılması gereken en önemli görevler hava araçlarının koldan çıkmalarına mani olmak ve hava araçlarının birbirlerine veya bir engele çarpmalarını önlemektir.

### 4.1. Kolda Kalabilme

Otonom kol uçuşu icra eden savaş uçakları arasında, yakıt ikmali gibi farklı görevlerde lider ve takipçiler arasında veya İHA'lar arasında kol düzeni/dizilimi korunurken istenen uçuş güzergâhına da yakınsamalı ve lider tarafından belirlen zamanla değişen yörünge kol tarafından takip edilebilmelidir. Kol dizilimi yaygın olarak üçgen seçilir. Üç hava aracı kolu oluşturur; lider öndedir ve iki takipçi, liderden alınan veriler aracılığıyla lideri takip eder. Bu üçgen düzeninin korunması ve herhangi bir hava aracının düzen dışına çıkmaması, başarılı kol uçuşu için kritik önemdedir.

Hava aracı otonom kol uçuşu için tasarlanacak kontrol sisteminde istenen uçuş güzergâhına göre hava araçlarını daha iyi kontrol etmek için, bir dış döngü ve bir iç döngünün eşzamanlı olarak yürütülmesi gerekir. Dış döngünün birincil amacı, her hava aracının yönelimini, ileri hızı ve konumu ile istenen yol arasındaki ilişkiyle başa çıkmaktır. Bu sırada iç döngü, dış döngüden komutlar alır ve ardından İHA'nın istenen irtifada iyi uçabilmesini sağlamak için kanatçık ve dümen komutlarını üretir. Kontrol sistemi bu amaçla hava araçlarının ivmeleri, üç eksen boyunca açısal ve çizgisel hızlarını, konum ve yönelim bilgilerini kullanır. İç döngü çoğu zaman hava aracı otopilot sistemleri ile sağlanırken dış döngü için MÖK kullanımı daha uygun ve yaygındır.

**4.2. Çarpışma veya Engelden Kaçınma**

Otonom kol uçuşu sırasında, hem hava araçları arasındaki hem de hava araçları ile engeller arasındaki çarpışmaları önlemek için, bir engelden kaçınma kontrol yöntemi özellikle önemlidir. Bu amaçla hava araçları arasındaki ve lider veya bir takipçi ile yakındaki bir engel arasındaki çarpışmaları önlemek için bir lider-takipçi stratejisi ile engelden kaçınma stratejisinin kol için ayrı ayrı tasarlanması gerekir.

Üçgen bir hava aracı kolunda, her hava aracı aynı ileri hıza ve yönelime sahip olmalıdır. Hava araçları sabit bir düzlemde bulunur ve böylece kol katı bir cisim olarak kabul edilip modellenebilir. Engeller ise şekillerine göre, örneğin, silindirik bir engel, bir daireye indirgenebilir. Böylece engellerden kaçınma problemi, çembere teğet olmayan bir yörüngede bir çember etrafında hareket eden bir cismin problemi şeklinde basitleştirilebilir. Engel ister statik ister dinamik olsun, engel ile hava araçları arasında çarpışmayı önlemek için cisim ile daire arasındaki ilişki teğet olmalı veya örtüşmemelidir. Çarpışmadan kaçınma algoritmaları koldaki hava araçları tarafından kontrolcü mimarisi dışında tutularak kullanılabileceği gibi MÖK mimarisi çarpışma riski oluşturan cisimleri ele alarak da hedef takip kontrolü yapabilir.

Çarpışmalardan kaçınma stratejisi olarak geliştirilmiş ve yaygın olarak kullanılan yöntemler mevcuttur. Bu yöntemler ele aldıkları engel yapılarına göre değişmektedir. Çarpışmadan kaçınırken engellere göre karşılaşılan sıkıntılar vardır. Örneğin bazı yöntemlerdeki problem yapısında engelin şekli dikdörtgen olarak sınırlandırılmışken, gerçek yaşam ortamında engellerin şekli herhangi bir biçimde olabilir. Her engeli dikdörtgen olarak kabul etmek, hava aracının gerekli manevrayı gerçekleştirmesi için faydalı olabilecek uçuş bölgelerini de kullanamamak anlamına gelebilir. Ayrıca çarpışmadan kaçınma kısıtlaması tahmin edilen gelecekteki konumlara dayalı olarak MÖK'te formüle edildiğinden, engelin genişliği hava aracının bir adımlık tahmini hareket mesafesinden daha küçükse (kontrol ufkunun uzun olması veya araç hızının yüksek olması nedeniyle olabilir), örneğin algoritma çarpışmaları önlemede başarısız olabilir. Bu sorunla ve engel şekillerinin çeşitliliğiyle başa çıkmak için de yöntemler geliştirilmelidir. Bu amaçla model öngörülü kontrolde kayan ufuklar, zamansal ve mekansal olmak üzere ikiye ayrılabilir. Zamansal ufuklar zaman aralıklarını yönetirken, mekansal ufuklar fiziksel alanları yönetebilir.

**Sonuç**

Bu bölümde kol uçuşu, otomatik kontrol ve MÖK ele alınarak kol uçuşunun otonom icrasında model öngörülü kontrol yaklaşımının sahip olduğu kapasite ortaya konulmuştur. MÖK'ün engellerden ve hava araçları arasındaki çarpışmalardan kaçınırken, istenen güzergahtan çok fazla sapmadan kol uçuşu gerçekleştirme yeteneğine sahip bir kontrolcü olduğu görülmektedir. Bu yöntemle üç boyutlu ortamda altı serbestlik derecesine sahip hava aracı modelleri daha çok üçgen diziliminde kol uçuşunu otonom gerçekleştirmiştir. MÖK mimarisi ile kol uçuşunun gerçekleşeceği güzergâh izlenebileceği gibi çarpışmadan kaçınma stratejileri de MÖK mimarisine gömülebilir. Yapılan benzetim çalışmaları her iki şekilde de MÖK'ün yeterli kabiliyete sahip olduğunu göstermektedir. Dolayısıyla bu yöntemin kısıtlamalarla başa çıkabildiği, hava araçları arası çarpışmaların yanı sıra engellerden de kaçınabildiği ve herhangi bir çarpışma riski olmadığında kol düzenini sağladığı görülmektedir.

Hava araçlarında model öngörülü kontrol çalışmalarının neredeyse tamamının benzetim ortamlarında gerçekleştirildiği görülmektedir. Bunun en önemli nedenleri MÖK'ün gerçek zamanlı uygulamalarda kullanımının yüksek işlem hacmi gerektirmesi gibi sahip olduğu zorluklardır. Havacılık dışındaki MÖK uygulamalarında ise bu zorluğun üstesinden gelebilmek için merkezi işlemci (CPU) yerine işlem hacmi daha büyük olan grafik işlemciler (GPU) kullanılabilmektedir. Dolayısıyla hava araçlarının sahip oldukları karmaşık hareket yapıları dikkate alınarak daha yüksek kapasiteli işlemcilerle MÖK uygulamalarının gerçek zamanlı yapılması üzerine çalışılması gerekmektedir.

Mühendislik Alanında Gelişmeler – 2